\documentclass[10pt,twocolumn,a4paper]{article}

\usepackage[a4paper, margin=1.8cm]{geometry}
\usepackage[utf8]{inputenc} % allow utf-8 input
\usepackage[T1]{fontenc}    % use 8-bit T1 fonts
\usepackage{times}
\usepackage[pagebackref,breaklinks,colorlinks]{hyperref}
\usepackage{authblk}
\usepackage{graphicx}
\usepackage{multirow}
\usepackage{amsmath,amssymb,amsfonts}
\usepackage{amsthm}
\usepackage{mathrsfs}
\usepackage[title]{appendix}
\usepackage{lineno}
\usepackage{xcolor}
\usepackage{textcomp}
\usepackage{manyfoot}
\usepackage{booktabs}
\usepackage{algorithm}
\usepackage{algorithmicx}
\usepackage{algpseudocode}
\usepackage{listings}
\usepackage{realboxes}
\usepackage{arydshln}
\usepackage{xspace}
\usepackage{orcidlink}
\usepackage{multirow}
\usepackage{placeins}

\newcommand{\ie}{\textit{i.e.}\xspace}
\newcommand{\eg}{\textit{e.g.}\xspace}
\newcommand{\Ie}{\textit{I.e.}\xspace}

\newcommand{\cf}{\textit{cf.}\xspace}
\newcommand{\etc}{\textit{etc.}\xspace}
\newcommand{\wrt}{\textit{w.r.t.}\xspace}

\title{A Guide to Structureless Visual Localization} 

\author[1,2]{Vojtech Panek\orcidlink{0000-0003-0601-7682}}
\author[3]{Qunjie Zhou\orcidlink{0000-0002-2434-2393}}
\author[4]{Yaqing Ding\orcidlink{0000-0002-7448-6686}}
\author[3]{Sérgio Agostinho\orcidlink{0000-0001-7008-1756}}
\author[4]{\\Zuzana Kukelova\orcidlink{0000-0002-1916-8829}}
\author[2]{Torsten Sattler\orcidlink{0000-0001-9760-4553}}
\author[3]{Laura Leal-Taixé\orcidlink{0000-0001-8709-1133}}

\affil[1]{Faculty of Electrical Engineering, Czech Technical University (CTU) in Prague}
\affil[2]{Czech Institute of Informatics, Robotics and Cybernetics, CTU in Prague}
\affil[3]{NVIDIA}
\affil[4]{Visual Recognition Group, Faculty of Electrical Engineering, CTU in Prague}

\date{}

\hypersetup{
pdftitle=[A Guide to Structureless Visual Localization]{A Guide to Structureless Visual Localization},
pdfsubject={cs.CV},
pdfauthor={Vojtech Panek, Qunjie Zhou, Yaqing Ding, Sérgio Agostinho, Zuzana Kukelova, Torsten Sattler, Laura Leal-Taixé},
pdfkeywords={Visual Localization, Pose Regression, Pose Triangulation},
}

\begin{document}
\maketitle

\begin{abstract}
    Visual localization algorithms, i.e., methods that estimate the camera pose of a query image in a known scene, are core components of many applications, including self-driving cars and augmented / mixed reality systems. State-of-the-art visual localization algorithms are structure-based, i.e., they store a 3D model of the scene and use 2D-3D correspondences between the query image and 3D points in the model for camera pose estimation. While such approaches are highly accurate, they are also rather inflexible when it comes to adjusting the underlying 3D model after changes in the scene. Structureless localization approaches represent the scene as a database of images with known poses and thus offer a much more flexible representation that can be easily updated by adding or removing images. Although there is a large amount of literature on structure-based approaches, there is significantly less work on structureless methods. Hence, this paper is dedicated to providing the, to the best of our knowledge, first comprehensive discussion and comparison of structureless methods. Extensive experiments show that approaches that use a higher degree of classical geometric reasoning generally achieve higher pose accuracy. In particular, approaches based on classical absolute or semi-generalized relative pose estimation outperform very recent methods based on pose regression by a wide margin. Compared with state-of-the-art structure-based approaches, the flexibility of structureless methods comes at the cost of (slightly) lower pose accuracy, indicating an interesting direction for future work.
\end{abstract}

\section{Introduction}\label{sec:introduction}
Visual localization is the task of predicting the position and orientation, \ie, the camera pose, from which a given image was taken. 
Solving the visual localization task is a core step in many applications, including autonomous robots such as self-driving cars and drones, and augmented / mixed reality applications~\cite{Heng2018ProjectAL,Lim2012RealtimeI6,Lynen2015GetOO, Middelberg2014Scalable6L, Panek2022ECCV}.

State-of-the-art visual localization algorithms use 2D-3D matches between pixels in a query image and 3D points in the scene to estimate the camera pose~\cite{sarlin2019coarse, Sattler2017EfficientE, Panek2022ECCV, Taira2019TPAMI, zhou2022gomatch, wang2024dgc}. 
These matches are typically either established by matching local feature descriptors~\cite{sarlin2019coarse,sarlin2020superglue,Sattler2017EfficientE,Panek2022ECCV,Panek2023VisualLU,Li2012WorldwidePE} or regressed directly via a neural network~\cite{Shotton2013SceneCR,Cavallari2019TPAMI,Brachmann2023AcceleratedCE,Brachmann2020VisualCR}. 
Given such 2D-3D matches, the camera pose can then be estimated by applying a perspective-n-point (PnP) pose solver~\cite{grunert1841pothenotische}, \eg, a P3P solver for calibrated cameras~\cite{haralick1994review,Fischler1981RandomSC,Persson2018ECCV,ding2023revisiting}, inside a RANSAC framework~\cite{Fischler1981RandomSC,Lebeda2012BMVC,chum2008optimal,Barth2019MAGSACAF}.
Alternatively, the camera pose can also be estimated by refining an initial pose estimate, \eg, obtained from image retrival~\cite{Jgou2010AggregatingLD, arandjelovic2013all,Arandjelovi2015NetVLADCA,GARL17,RARS19,Berton_2023_EigenPlaces}, via a render-and-compare approach~\cite{sarlin21pixloc, Pietrantoni_2023_CVPR, pietrantoni2024self, trivigno2024unreasonable, Lin2020iNeRFIN, zhou2024nerfect, chen2022dfnet, chen2023nefes, liu2023nerfloc, sun2023icomma, botashev2024gsloc, zeller2024gsplatloc, liu2024gs}: 
Given a current pose estimate, the query image is compared to a rendering of the scene (either in the form of colors or by rendering a feature space). 
The pose is then optimized as to increase the consistency between the actual image and the rendering. 

Common to these approaches is that they represent scenes via 3D models, either storing the models explicitly, \eg, in the form of Structure-from-Motion (SfM) point clouds, meshes, or 3D Gaussian Splats~\cite{kerbl20233d}, or implicitly, \eg, in the weights of a neural network or a neural radiance field (NeRF)~\cite{Mildenhall2020NeRFRS}. 
While 3D model-based scene representations facilitate highly accurate camera pose estimation, they are also rather inflexible. 
In the case of changes in the scene, \eg, a newly constructed building, a building being renovated, furniture being moved or replaced, \etc, detecting the changes~\cite{yew2021city,adam2022objects} and updating the 3D model accordingly are typically complex tasks by themselves~\cite{Torii2019TPAMI}. 

An alternative to the structure-based approaches discussed above are structureless visual localization methods~\cite{Balntas2018ECCV,Zhang2006ImageBL,Laskar2017ICCVW,Bhayani2021CalibratedAP,Zhou2020ICRA,dong2023lazy, arnold2022map}. 
Structureless approaches represent the scene through a database of images, each associated with a camera pose and camera intrinsics. 
Given a query image, they first use image retrieval to identify a set of relevant database images. 
They then estimate the camera pose of the query relative to the poses of the top-retrieved images~\cite{Zhou2020ICRA, arnold2022map,Kazhdan2013TOG,Bhayani2021CalibratedAP,Zheng_2015_ICCV}. 
Since each database image is treated independently of the other, adding images with new observations or deleting images that show outdated versions of a scene is trivial.\footnote{It might even not be necessary to delete images at all as outdated images will eventually not be retrieved or will not provide information when computing the pose of the query.}  

Structureless methods were among the first visual localization approaches~\cite{Zhang2006ImageBL}. 
However, there is significantly less work on structureless methods than on structure-based methods. 
Interestingly, a large part of the literature on structure-based methods does not compare with structureless approaches. 
At the same time, to the best of our knowledge, there is not even a comprehensive comparison of structureless methods between themselves. 
This paper aims to close this gap in the literature.

In detail, this paper makes the following contribution: 
(\textbf{1}) we provide a comprehensive review of structureless localization methods. 
(\textbf{2}) through extensive experiments, we compare state-of-the-art versions of the most important families of structureless methods: 
pose triangulation, semi-generalized relative pose estimation, absolute pose estimation via locally triangulated 3D point clouds, and relative pose regression. 
For each method, we first ablate multiple variants (mostly by evaluating the use of different local features). 
We then compare between the different families using the best-performing version from the previous experiments. 
(\textbf{3}) the results of our experiments lead to multiple interesting insights: 
(\textit{a}) 
More explicit use of geometric reasoning often leads to better pose accuracy. 
\Ie, methods relying on pose triangulation from relative poses perform worse than methods based on semi-generalized relative pose estimation (which estimates the pose of the query simultaneously \wrt multiple database images instead of computing only pairwise relative poses). 
In turn, methods based on locally triangulating the 3D scene structure followed by absolute pose estimation can provide more accurate pose predictions than approaches based on semi-generalized relative pose estimation. 
Interestingly, relative pose regression-based approaches perform worst, despite strong recent progress~\cite{dust3r_cvpr24,MASt3R_eccv24, dong2024reloc3r}.
(\textit{c}) In terms of their pose accuracy-run-time trade-off, methods using less geometric reasoning can offer a better performance, \ie, methods that do not provide the highest pose accuracy are still useful in practice. 
(\textit{d}) There is no type of  features that performs best in all scenarios. 
The best type of feature to use depends on the method and dataset.
(\textbf{4}) We provide a comparison of the best-performing structureless methods with state-of-the-art structure-based approaches. 
Our results show that structureless approaches can be competitive with structure-based methods. 
Thus, improving the accuracy of structureless localization algorithms is an interesting direction for future work.

\section{Related Work}
As this paper focuses on comparing structureless approaches for visual localization, we focus our discussion on such methods. For completeness, we also review structure-based approaches. 

\subsection{Place Recognition}
Highly related to the (structureless) visual localization problem is the task of visual place recognition~\cite{Baatz10ECCV,Baatz2011IJCV,Torii-CVPR15,Chen2011CVPR,Chen2017ICRA,lowry2016visual,Hausler_2021_CVPR,Berton_2023_EigenPlaces,Berton_CVPR_2022_CosPlace,Arandjelovic2014ACCV,Singh2016LSVGL,Zamir14PAMI,Ardeshir2014ECCV,Zamir10ECCV,Cao13CVPR}. 
Given a query image and a database of geo-tagged photos, the goal of place recognition is to identify the scene depicted in the query. 
This is typically done by retrieving a database image depicting the same place. 
Place recognition approaches thus build on image retrieval techniques~\cite{Arandjelovi2015NetVLADCA,Torii-CVPR15,Berton_2023_EigenPlaces,Berton_CVPR_2022_CosPlace,Radenovic2019PAMI}.  
The classical image retrieval task aims to identify all images depicting the same content as the query photo~\cite{Sivic03ICCV,Philbin07CVPR,Philbin10ECCV,Tolias2016IJCV}. 
In contrast, place recognition requires only a single relevant image among the top-n retrieved images~\cite{Torii-CVPR15}.  
In the context of structureless visual localization, place recognition approaches can be considered as pose approximation methods, which approximate the pose of the query by the poses of the top-retrieved images.

\subsection{Structure-based Localization}

\noindent \textbf{Feature matching-based.} 
Early visual localization approaches based on feature matching represented the scene via Structure-from-Motion point clouds~\cite{Se2002IROS,Li2010ECCV,Li2012WorldwidePE,Sattler2011ICCV,Sattler2012ECCV,Choudhary12ECCV,Irschara09CVPR,Arth09ISMAR,Sattler2012BMVC}. 
Each 3D point was triangulated from features extracted from the database images used to build the SfM model. 
Thus, each point can be associated with the descriptors of its corresponding local image features. 
As a result, 2D-3D matches can be established by comparing feature descriptors extracted from the query image to descriptors associated with the 3D points~\cite{Se2002IROS,Li2010ECCV,Li2012WorldwidePE,Sattler2011ICCV,Sattler2012ECCV,Choudhary12ECCV}. 
The resulting 2D-3D correspondences are then used to estimate the camera pose of the query image by applying a PnP solver, typically a P3P solver for calibrated cameras~\cite{haralick1994review,Fischler1981RandomSC,Persson2018ECCV,ding2023revisiting}, inside  RANSAC~\cite{Fischler1981RandomSC,Lebeda2012BMVC,chum2008optimal,Barth2019MAGSACAF}. 

While such methods based on directly matching query and 3D point descriptors can run on mobile devices~\cite{Arth09ISMAR,Lynen2015GetOO}, they struggle to scale to larger scenes~\cite{Li2012WorldwidePE}. 
This is due to ambiguities in local appearance arising in larger and more complicated scenes: 
At scale, it is likely that there are many 3D points with similar descriptors, making it hard to identify the corresponding 3D point for a given query feature~\cite{Li2012WorldwidePE}. 
These ambiguities can be resolved either by accepting more incorrect matches and filtering them out through geometric reasoning~\cite{Svarm2017PAMI,Zeisl2015ICCV}, or by using image retrieval to restrict matching to subparts of the 3D model~\cite{Irschara09CVPR,Sattler2012BMVC,Sattler2015ICCV,Taira2019InLocIV,HumenbergerX20Kapture,Sarlin2018CORL,sarlin2019coarse}. 
The latter methods are known as hierarchical localization approaches, which only match query features against 3D points visible in the top-retrieved database images and constitute the current state of the art. 

Although SfM point clouds are the dominant scene representation for feature-based methods, other representations, including  meshes~\cite{Panek2022ECCV,Panek2023CVPR}, Neural Radiance or Feature Fields~\cite{liu2023nerfloc,chen2023nefes,zhou2024nerfect}, and 3D Gaussian Splats~\cite{Matteo2024ECCV}, are used. 
Still, these approaches are based on establishing 2D-3D matches using local features.

\noindent \textbf{Scene coordinate regression.} 
Rather than relying on descriptor matching to establish 2D-3D correspondences, scene coordinate regression methods train machine learning models to directly regress the corresponding 3D point position for a given input patch~\cite{Brachmann2017CVPR, Brachmann2020VisualCR, Brachmann2018CVPR, li2020hierarchical, dong2022src, tang2021dsm,Shotton2013SceneCR,Cavallari2019TPAMI,Brachmann2023AcceleratedCE,Cavallari20193DV,wang2023hscnet,Budvytis2019BMVC,Massiceti2017ICRA,Guzman2014CVPR,Valentin2015CVPR}. 
While early approaches used random forests~\cite{Shotton2013SceneCR,Massiceti2017ICRA,Valentin2015CVPR,Guzman2014CVPR}, recent methods use neural networks. 
As for feature-based methods, the resulting 2D-3D correspondences are then used for RANSAC-based pose estimation. 
Whether scene coordinate regressors or feature-based approaches are more accurate is still an open question~\cite{Brachmann2021OnTL}.

\noindent \textbf{Camera pose regression relative to 3D models.} 
Both feature-based methods and scene coordinate regressors establish 2D-3D matches for pose estimation. 
Given that an initial coarse pose estimate is typically available, \eg, obtained via image retrieval, an alternative approach to structure-based localization is pose refinement. 
Such methods iteratively compare the query image against a rendering of a 3D model from the current pose estimate~\cite{sarlin21pixloc, Pietrantoni_2023_CVPR, pietrantoni2024self, trivigno2024unreasonable, Lin2020iNeRFIN, chen2023nefes, sun2023icomma, botashev2024gsloc, zeller2024gsplatloc, liu2024gs,VonSturmberg20203DV,VonSturmberg2020RAL}. 
The pose is then adjusted to reduce the difference between the image and the rendering. 
In terms of pose accuracy and robustness, such regression approaches are inferior to hierarchical feature-based methods. 

\subsection{Structureless Localization}
\noindent \textbf{Pose approximation.} Given a database of images with known camera poses, the pose of a query image can be approximated efficiently through the pose of the top-retrieved image~\cite{Torii-CVPR15}. 
Better approximations can be obtained by interpolating the poses of the top-n retrieved images~\cite{Torii-CVPR15,Sattler2019CVPR}. 
\cite{Thoma2020RAL} proposed to learn descriptors for image retrieval such that distances in the descriptor space are proportional to pose similarity. 
The resulting descriptors enable more accurate pose approximation. 
Still, the pose quality is significantly below the requirements of precise visual localization. 
Hence, we do not evaluate pose approximation approaches in this work. 

\noindent \textbf{Pose triangulation.} 
A more precise query pose can be obtained from the relative poses between the query and the top-retrieved database images~\cite{Zhang2006ImageBL,Laskar2017ICCVW,Zhou2020ICRA,dong2023lazy}. 
Pairwise relative poses, \eg, computed by estimating an essential matrix or homography, only provide the direction of the relative translation, but not its magnitude. 
Given that the poses of the database images are known, the position of the query photo can be triangulated from two or more relative translation directions~\cite{Zhang2006ImageBL,Laskar2017ICCVW,Zhou2020ICRA}. 
Inspired by global SfM approaches~\cite{pan2024global, cui2015global, zhu2018very}, LazyLoc~\cite{dong2023lazy} further improves pose accuracy by adding rotation and translation averaging stages followed by a post optimization that jointly optimizes the query camera pose and the 3D points triangulated from 2D feature tracks.
In this work, we consider both a "standard" pose triangulation approach~\cite{Zhou2020ICRA} and LazyLoc~\cite{dong2023lazy}. 

\noindent \textbf{Semi-generalized relative pose estimation.} 
The scale of the translation can be recovered when computing the pose of the query \wrt two or more database photos~\cite{Zheng_2015_ICCV}, \ie, when computing a semi-generalized relative pose rather than the relative pose between two images. 
Given the known poses of the database images, the relative pose of the query can then be directly translated into its absolute pose. 
\cite{Zheng_2015_ICCV} derive multiple solvers for computing the semi-generalized relative pose between a query and two database images, where the relative pose between the database photos is known. 
The solvers differ based on the number of correspondences between the query and the other two images. 
However, all of the solvers except one are too slow to be used in practice. 
The remaining solver assumes 5 point correspondences between the query and one of the database images and a single correspondence between the query and the other database image. 
The 5 matches are used to compute the relative pose by estimating the essential matrix, which can be done highly efficiently~\cite{Nister-5pt-PAMI-2004}. 
The remaining correspondence is then used to estimate the scale of the translation. 
The solver is known as the E5+1 solver. 
We use a method that applies the solver inside a RANSAC loop as one of our baselines. 
As our experiments show, this approach is very competitive. 
Interestingly, it has not been used as a baseline in other works on structureless localization such as~\cite{Laskar2017ICCVW,dong2023lazy,Zhou2020ICRA}. 

\cite{Bhayani2021CalibratedAP} shows that more efficient solvers for the setting described above can be derived by assuming that the scene is locally planar. 
These solvers are based on estimating homographies, hence \cite{Bhayani2021CalibratedAP} solve a semi-generalized homography estimation problem. 
In our experiments, we only evaluate a method build around the E5+1 solver due to its efficiency and easy implementation.

\noindent \textbf{Constructing SfM models on the fly.} 
A common approach to pose triangulation and semi-generalized relative pose estimation is to establish 2D-2D correspondences between the query and the top-retrieved database images. 
Implicitly, these matches define point correspondences between the retrieved database images. 
Together with the known poses of the database images, these correspondences can be used to triangulate 3D points. 
This results in 2D-3D matches for the query, which can then be used for pose estimation. 
As our experiments show, approaches that build such local SfM models on the fly~\cite{Torii2019TPAMI,Pion20203DV,Humenberger2022IJCV} lead to the best pose accuracy among all tested structureless localization strategies. 
Their downside is the computational overhead caused by triangulation. 

\noindent \textbf{Absolute pose regression.} 
The structureless localization approaches discussed above, with the exception of pose approximation methods, all establish 2D-2D feature matches between the query and the database images. 
These matches in turn are used to explicitly estimate the geometric relation between the photos. 
An alternative approach is to train a neural network to directly regress the pose of the query image~\cite{Kendall2015ICCV,Kendall2017CVPR,Walch2017ICCV,Moreau2021CORL,Shavit2021ICCV,Brahmbhatt2018CVPR}. 
However, as shown in~\cite{Sattler2019CVPR}, most of these absolute pose regression approaches are not significantly more accurate than pose approximation methods. 
These methods implicitly store the scene through the weights of a neural network. 
As such, updating the scene representation is non-trivial and will require fine-tuning the network. 
For these reasons, we do not consider absolute pose regression approaches in our experimental comparison. 

\noindent \textbf{Relative pose regression.} 
Rather than regressing the absolute pose of a single image, relative pose regression approaches regress the relative pose between two input images (potentially including the scale of the translation)~\cite{Laskar2017ICCVW,Zhou2020ICRA,Balntas2018ECCV,Ng20223DV,dust3r_cvpr24,MASt3R_eccv24,Ding2019ICCV,dong2024reloc3r,wang2025vggt}. 
In the context of structureless localization, relative pose regressors can thus be used instead of explicit geometric reasoning from feature matches. 
Although initial approaches were not significantly better than pose approximation algorithms~\cite{Sattler2019CVPR}, recent methods~\cite{MASt3R_eccv24,dust3r_cvpr24,dong2024reloc3r,wang2025vggt} are at least competitive with classical approaches. 
Under challenging conditions, especially if there is little visual overlap between images, they can significantly outperform classical approaches. 
Thus, we include the recent pose regression-based approaches~\cite{MASt3R_eccv24,dong2024reloc3r} in our evaluation. 

\section{Selected Methods}
This paper aims at understanding the performance of existing structureless visual localization approaches by comparing them through extensive experiments. 
In the following, we discuss the methods we selected for our comparison, grouped based on the family of approaches they belong to. 

\noindent \textbf{Pose triangulation.} 
We evaluate two pose triangulation approaches: 
\emph{Localization from essential matrices}~\cite{Zhou2020ICRA} (\emph{Ess. mat.}) computes the relative poses between the query and the retrieved database images using the well-known 5-point algorithm~\cite{Nister-5pt-PAMI-2004} (further dubbed as 5Pt) inside a RANSAC~\cite{Fischler1981RandomSC,Lebeda2012BMVC} loop.
The final orientation of the query image is computed via averaging the relative rotations. 
The camera position is estimated by triangulation using the estimated translation directions.
To evaluate this method, we reimplemented the functionality of the code provided with the original publication~\cite{Zhou2020ICRA}.

We further evaluate a variant of \emph{Ess. mat.} that uses a 3-point solver~\cite{Ding2025RePoseDER} instead of the 5-point solver to compute relative poses. 
The 3-point solver uses monocular depth predictions to compute the relative pose from fewer 2D-2D matches, making it more suitable for scenarios with low inlier ratios. 
The 3-point solver is similarly accurate as the 5-point solver when using reasonably accurate depth maps~\cite{Ding2025RePoseDER}. 
We denote the resulting approach as \emph{Ess. mat. (3Pt+depth)}. 
We use the solver implementation of the authors of~\cite{Ding2025RePoseDER} together with our reimplementation of the \emph{Ess. mat.} method.

\emph{LazyLoc}~\cite{dong2023lazy} also uses the 5-point algorithm~\cite{Nister-5pt-PAMI-2004} to obtain relative poses between the retrieved and the query image. 
The query pose is then computed using robust motion averaging with outlier rejection, followed by query pose optimization based on 2D-3D matches. 
For the latter, 3D points are triangulated from 2D-2D matches between the retrieved database images. 
The pose is then optimized via least-squares minimization of reprojection errors. 
The used implementation was kindly provided by the authors of~\cite{dong2023lazy}. 

We selected \emph{Ess. mat.} as an example of a rather straightforward pose triangulation approach, while \emph{LazyLoc} represents the current state-of-the-art in terms of pose triangulation. 

\noindent \textbf{Semi-generalized relative pose estimation.} 
As mentioned above, we build a localization system around the E5+1 solver~\cite{Zheng_2015_ICCV}. 
The solver first estimates relative pose between the query image and one database image from 5 2D-2D correspondences using the 5-point solver~\cite{Nistr2004AnES}. 
One additional 2D-2D match between the query and another retrieved image is then used to recover the scale of the translation based on the known poses of the two database images. 
We apply the solver within RANSAC with local optimization~\cite{Lebeda2012BMVC}. 
Local optimization is performed by least squares minimization of Sampson errors~\cite{Hartley2001MultipleVG} starting from poses estimated by the E5+1 solver. 
We use the implementation provided by PoseLib~\cite{PoseLib} and refer to the approach as \emph{E5+1}. 

As for \emph{Ess. mat.}, we also evaluate a variant of \emph{E5+1} that uses the 3-point solver from~\cite{Ding2025RePoseDER} instead of the 5-point solver to compute relative poses. 
We denote this approach as \emph{E3+1}. 

\noindent \textbf{SfM on the fly.} 
We evaluate two variants of an SfM on-the-fly pipeline. 
Both use 2D-2D matches between the query and the database images to obtain 2D-2D correspondences between the database photos.\footnote{We do not match features between database images themselves.} 
In both cases, these 2D-2D matches are used to triangulate 3D points. 
In turn, these points define 2D-3D matches for the query image, which are then used for absolute pose estimation with a P3P solver~\cite{ding2023revisiting} inside RANSAC with local optimization.
Given that we build local 3D models on demand, \ie, based on the retrieved database images, and do not construct and maintain a single global model, we refer to both methods as \emph{Local triangulation}. 

The first variant uses all the database images retrieved for the single query image for triangulation (and is hence denoted as \emph{Local triangulation - all}). 
Each feature in a query image defines a track containing feature keypoints found in the retrieved database images. 
Triangulation is then performed for every keypoint in the track within RANSAC, selecting the 3D point with the highest number of inliers in terms of a given reprojection error threshold. 

The second variant, denoted as \emph{Local triangulation - pairs}, considers pairs of database images for triangulation and pose estimation. 
For each potential pair of retrieved database images, it triangulates 3D points and estimates the query pose from the resulting 2D-3D matches. 
The final query pose is the one with the largest number of inliers. 

For our experiments, we use both features based on sparse keypoint detections~\cite{DeTone2017SuperPointSI,Zhao2023ALIKED, Zhao2022ALIKE} and dense feature matchers~\cite{edstedt2024roma,MASt3R_eccv24}. 
The latter obtain 2D-2D correspondences by matching densely extracted features between two images. 
As a result, there are no repeatable keypoint detections between image pairs~\cite{Zhou2020Patch2PixEP,Sun2021LoFTRDL}.  
In order to be able to form tracks for dense matchers, matches having keypoints with mutually nearest coordinates in the query image (up to a distance threshold) are established as matches between the reference images. 
We used a distance threshold of 5 px in our experiments selected based on grid search. 
We implemented the pipelines using the point triangulation method from OpenCV~\cite{opencv_library} and the P3P solver from PoseLib~\cite{PoseLib, Persson2018ECCV}.

\noindent \textbf{Relative pose regression.} 
We evaluate multiple approaches based on relative pose regression~\cite{MASt3R_eccv24,dong2024reloc3r}. 
The first method is built around the MASt3R~\cite{MASt3R_eccv24} foundation model. 
Given the query and the retrieved database images, we use MASt3R to build a local 3D model. 
To this end, MASt3R first regresses pairwise depth maps (essentially using the DUSt3R~\cite{dust3r_cvpr24} approach), which are subsequently aligned, followed by optimizing the resulting camera poses. 
We then align this local reconstruction with the known poses of the database images using the Kabsch-Umeyama alignment~\cite{kabsch1976solution, umeyama1991least} in two stages: the first stage only uses the camera positions for alignment, thus aligning the camera positions and recovering the scale of the local reconstruction. 
In the case that the camera centers of the database images are (nearly) collinear, the alignment is ambiguous, as it is defined up to a rotation around the line containing the camera centers.
We thus add a single 3D point on the optical axis of every reference image, one length unit from the camera center. 
We then recompute the alignment using the database image positions and the additional points.\footnote{Note that the initial alignment stage is needed to recover the scale of the local reconstruction, which is necessary for computing the additional points.}
During our experiments, we observed that if there is a wrongly retrieved reference image without any overlap with query or other retrieved images, the optimizer still generates an arbitrary pose for it.
As the wrong pose can then skew the global alignment, we want to prevent the use of such images.
Therefore, the whole process of regression, optimization, and alignment is performed using a randomly sampled subset of the retrieved reference images and repeated multiple times.
We also filter out all retrieved images that don't have enough correspondences with query.
The final pose estimate is selected on the basis of the number of inliers in terms of a given epipolar error threshold.
We use the depth map regressor, the matcher, and the camera optimizer implemented within the MASt3R codebase together with our implementation of the alignment of the camera pose to the world frame.
We denote this MASt3R-based localization approach as \emph{MASt3R pose align}.

Note that the \emph{MASt3R pose align} approach is suboptimal in the sense that the MASt3R implementation released by the authors of~\cite{MASt3R_eccv24} cannot make use of known intrinsics and camera poses of the database images. 
However, to the best of our knowledge, the same limitation applies to all other 3D reconstruction approaches based on the relative pose regression~\cite{dust3r_cvpr24,wang2025vggt,dong2024reloc3r}. 
In all cases, adjusting the implementation is highly non-trivial, and we did not attempt the modification. 
Rather, we see the \emph{MASt3R pose align} approach as a way to measure how well existing relative pose-based approaches work out of the box.

Based on the MASt3R paper~\cite{MASt3R_eccv24}, the depth regressor was trained to perform metric-scale predictions.
If the depth maps are metric, they can be used to lift the 2D-2D correspondences to 3D.
The resulting 2D-3D matches can be used for pose estimation with P3P~\cite{Persson2018ECCV}.
We refer to this method as \emph{MASt3R depth + P3P}.

Another approach we tested is using the pairwise relative poses generated by MASt3R with the depth map alignment and estimating the query camera pose using the \emph{Ess. mat.} method.
This approach is further referred to as \emph{Ess. mat. (MASt3R poses)}.

The last approach we evaluate is \emph{Reloc3r}~\cite{dong2024reloc3r}. 
\emph{Reloc3r} uses a neural network to predict relative camera poses between query-database image pairs. 
The absolute pose of the query image is then obtained via pose triangulation:
The query image's orientation is obtained by averaging the rotation matrices predicted from the relative poses to the database images and the known poses of the database images. 
The query's camera position is computed using triangulation via the relative translation directions. 
The evaluation was performed using the implementation provided by the authors of the paper.

\section{Experimental Evaluation}\label{sec:experimental_evaluation}

\begin{figure*}[!tb]
    \centering
    \includegraphics[width=0.9\linewidth]{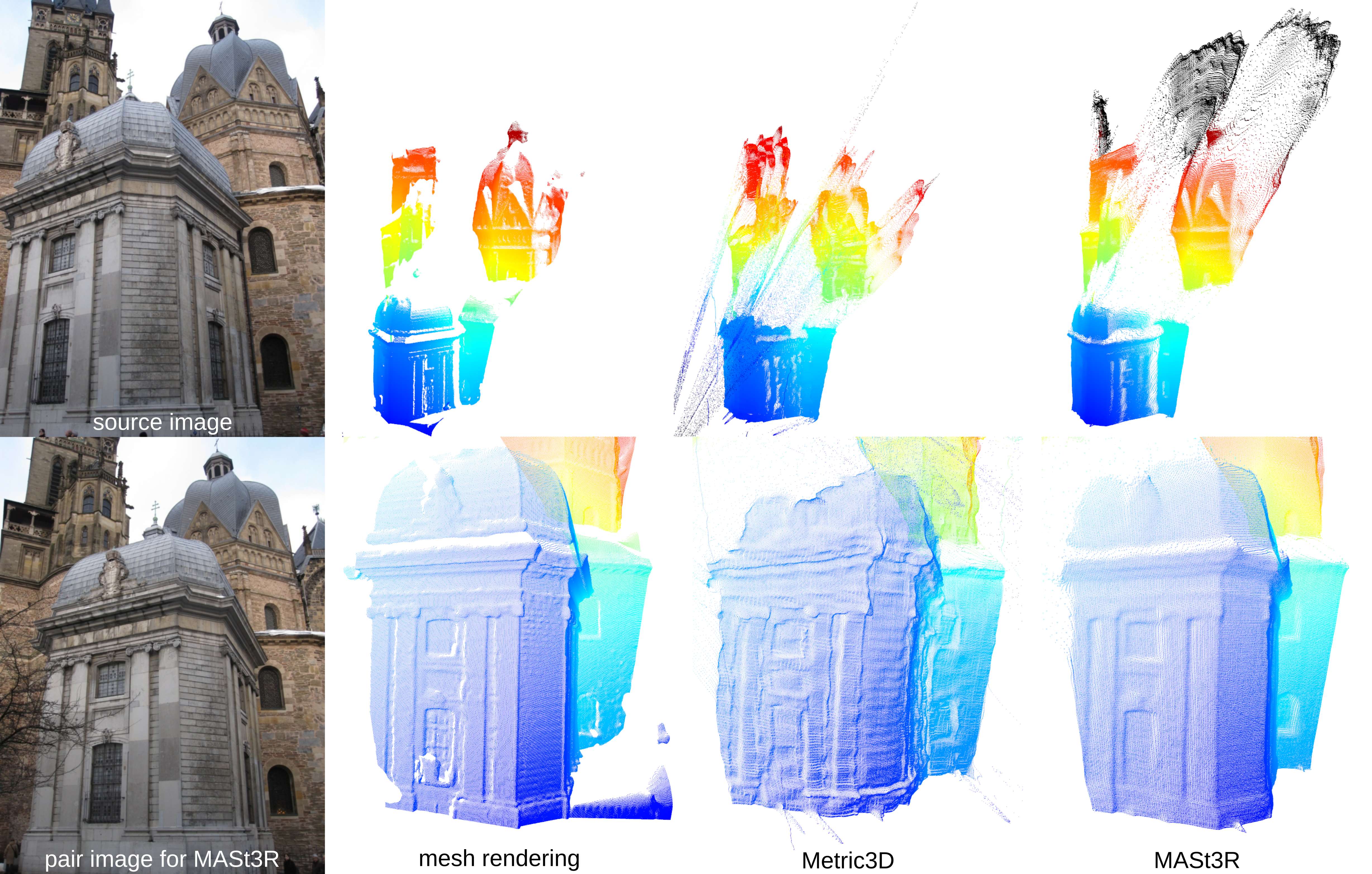}
    \caption{Comparison of depth maps from different sources - on the top left is the source image. The corresponding source camera was used for the rendering of a mesh model (AC-14 model from MeshLoc~\cite{Panek2022ECCV}). The source image together with its focal length is the sole input into the Metric3D v2~\cite{yin2023metric, Hu2024Metric3DVA} monocular depth estimator. As MASt3R~\cite{dust3r_cvpr24, MASt3R_eccv24} is a stereo model, it also uses a second image (shown in the bottom left) to predict the 3D geometry. MASt3R performs the prediction without any knowledge about the camera parameters. Both Metric3D and MASt3R depth maps were aligned (in scale and shift) to the mesh depth map for easier comparability, while they are used in their raw unscaled form in the experiments.}
    \label{fig:depth_maps}
\end{figure*}

\begin{figure*}[!tb]
    \centering
    \includegraphics[width=1\linewidth, trim=0 0.1cm 0 0.1cm, clip]{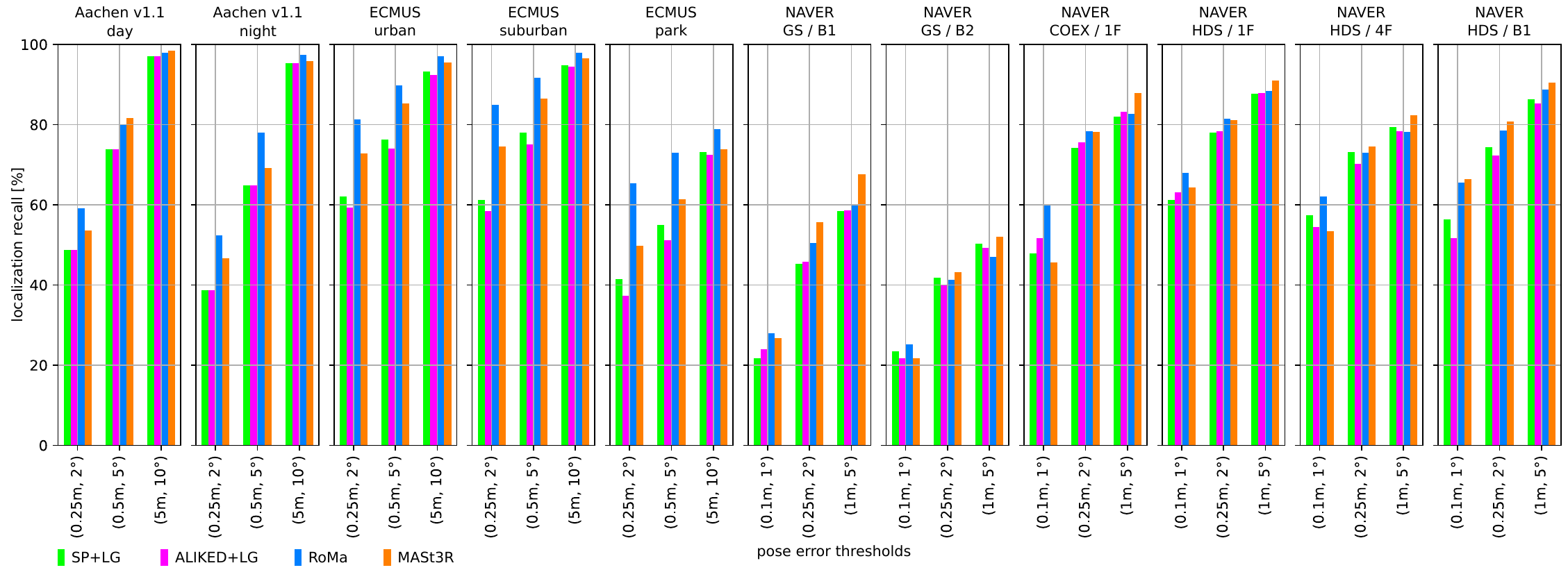}
    \vspace{-0.5cm}
    \caption{Localization results for the \emph{Ess. mat. (5Pt)} approach for different features. We report localization recalls (higher is better) on the Y-axis at multiple pose thresholds (X-axis). For the outdoor scenes, the best results are obtained with the RoMa matcher. For the indoor scenes, the MASt3R matcher performs best for the coarser thresholds.}
    \label{fig:recalls_rel_pose_triang}
\end{figure*}

\noindent \textbf{Datasets.} 
We evaluate the structureless visual localization approaches discussed above on multiple large datasets commonly used to benchmark visual localization algorithms:
Aachen Day-Night v1.1~\cite{Zhang2020ARXIV,Sattler2018CVPR, Sattler2012BMVC} is an outdoor dataset that captures the historic center of Aachen, containing time of day and seasonal changes.
Extended CMU Seasons~\cite{Sattler2018CVPR, Badino2011} is an outdoor dataset with seasonal changes with multiple urban, suburban, and park scenes captured from a moving car.
NAVER LABS Large-scale Localization Datasets in Crowded Indoor Spaces~\cite{Lee2021LargescaleLD} (hereinafter referred to as NAVER datasets) consist of multiple scenes in shopping malls and a large metro station.

\noindent \textbf{Evaluation protocol.} 
We follow common practice for the datasets and report the percentage of query images that are localized within specific error thresholds~\cite{Sattler2018CVPR}. 
\Ie, we report the percentage of query images localized within a $X$cm position error and a $Y^\circ$ rotation error of the corresponding ground-truth poses. 
We use the \href{https://www.visuallocalization.net/}{Long-Term Visual Localization benchmark} website~\cite{Sattler2018CVPR} for obtaining the measures.

\noindent \textbf{Implementation details.}
All methods assume that a set of reference images with known poses and camera intrinsics is given.
All baselines perform an initial image retrieval step using an image-level descriptor. 
Based on prior experience and preliminary experiments, we use the learned EigenPlaces descriptor~\cite{Berton_2023_EigenPlaces}.
Most of the selected methods use 2D-2D feature matches between the query and the retrieved set of reference images to compute the pose.
We evaluate two sparse local feature extractors, namely SuperPoint~\cite{DeTone2017SuperPointSI} and ALIKED~\cite{Zhao2023ALIKED, Zhao2022ALIKE}, in combination with the LightGlue~\cite{lindenberger2023lightglue} matcher. 
In addition, we also use two dense matchers, namely RoMa~\cite{edstedt2024roma} and MASt3R~\cite{dust3r_cvpr24, MASt3R_eccv24}. 
For RoMa, we use the "outdoor" model for the Aachen Day-Night v1.1 and Extended CMU Seasons datasets and the "indoor" model for the NAVER datasets.
In case of matching with MASt3R, we use only the coarse matching stage to keep the evaluation time within a reasonable scale.

For the \emph{Local triangulation} methods, we use a 2px reprojection error threshold for SuperPoint and ALIKED features, and a 8px threshold for RoMa and MASt3R.

For \emph{MASt3R pose align}, we sample a subset of 3 retrieved reference images and iterate the pose estimation 10 times.
Only reference images with more than 50 correspondences with the query are used, as a wrongly retrieved image without an overlap with the query can significantly skew the resulting pose estimate.
We use the two-stage optimization implemented in MASt3R~\cite{MASt3R_eccv24}.
The first stage minimizes 3D point map distances while optimizing only the cameras. 
The second stage minimizes reprojection errors while also changing the point maps.
We use 300 iterations for each stage, and learning rates of 0.2, respectively, 0.02 for the first, respectively, second stage.

For inlier counting, we use reprojection and epipolar error thresholds of 12 px.
For the \emph{E5+1}, \emph{E3+1} and local triangulation methods, we use a locally optimized RANSAC with a minimum of 1000 and a maximum of 100,000 iterations.

We use two state-of-the-art methods for depth prediction. 
The monocular metric depth estimator Metric3D v2~\cite{yin2023metric,Hu2024Metric3DVA} and the stereo geometry regressor from MASt3R~\cite{dust3r_cvpr24}.
We show a qualitative sample of the generated depth maps in Fig.~\ref{fig:depth_maps}.
Both estimated depth maps were aligned in global scale and shift to a depth map rendered from a mesh.\footnote{We used the AC-14 mesh model of the Aachen Day-Night v1.1 scene~\cite{Zhang2020ARXIV, Sattler2018CVPR, Sattler2012BMVC} provided by MeshLoc~\cite{Panek2022ECCV}. The mesh is metric, in the sense that distances can be measured in meters.}
Although the Metric3D depth map is not very accurate in detail and contains upsampling artifacts on depth discontinuities, it is able to recover the global scale relatively well.
For the presented image, the monocular depth was scaled down by a factor of 0.85 and shifted by +1.54 m during the alignment.
The MASt3R model is capable of reconstructing the geometry in remarkable detail, but its ability to estimate the scale of the scene is very poor.
The presented sample was scaled up by a factor of 15.83 and shifted by -0.10 m.
The two depth predictors performed similarly for other images from the Aachen v1.1 dataset.

\subsection{Ablation Studies} 

\begin{figure*}[!tb]
    \centering 
    \includegraphics[width=1.0\linewidth, trim=0 0.1cm 0 0.1cm, clip]{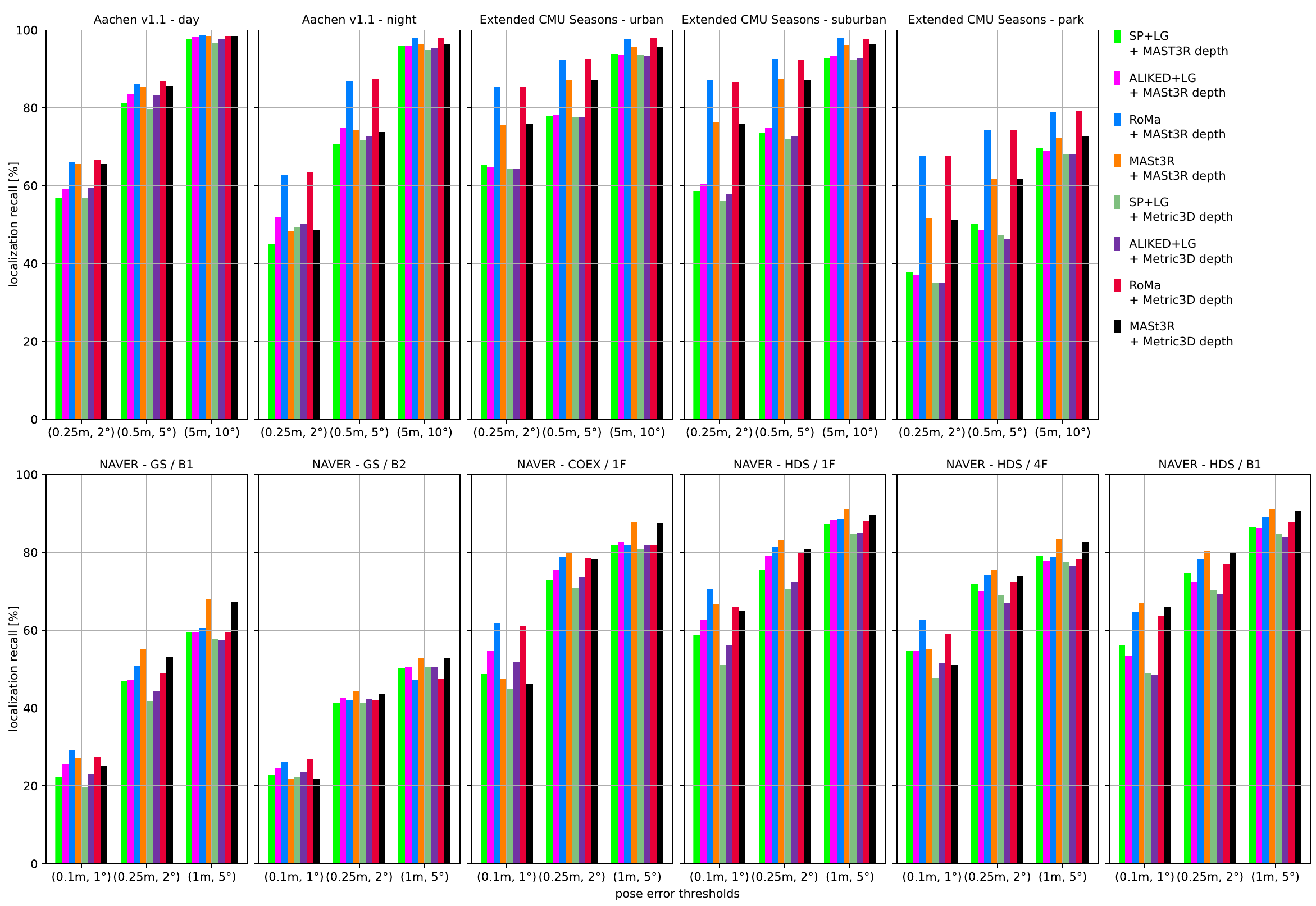}
    \vspace{-0.5cm}
    \caption{Localization results for the \emph{Ess. mat. (3Pt + depth)} approach for different  features and monocular depth predictors. We report localization recalls (higher is better) on the Y-axis at multiple pose thresholds (X-axis). For most scenes, the choice of the depth predictor is not critical. For outdoor scenes, RoMa yields the best results. For indoor scenes, MASt3R leads to the highest pose accuracy in most cases.}
    \label{fig:recalls_depth_relpose}
\end{figure*}

\begin{figure*}[!tb]
    \centering
    \includegraphics[width=1\linewidth, trim=0 0.1cm 0 0.1cm, clip]{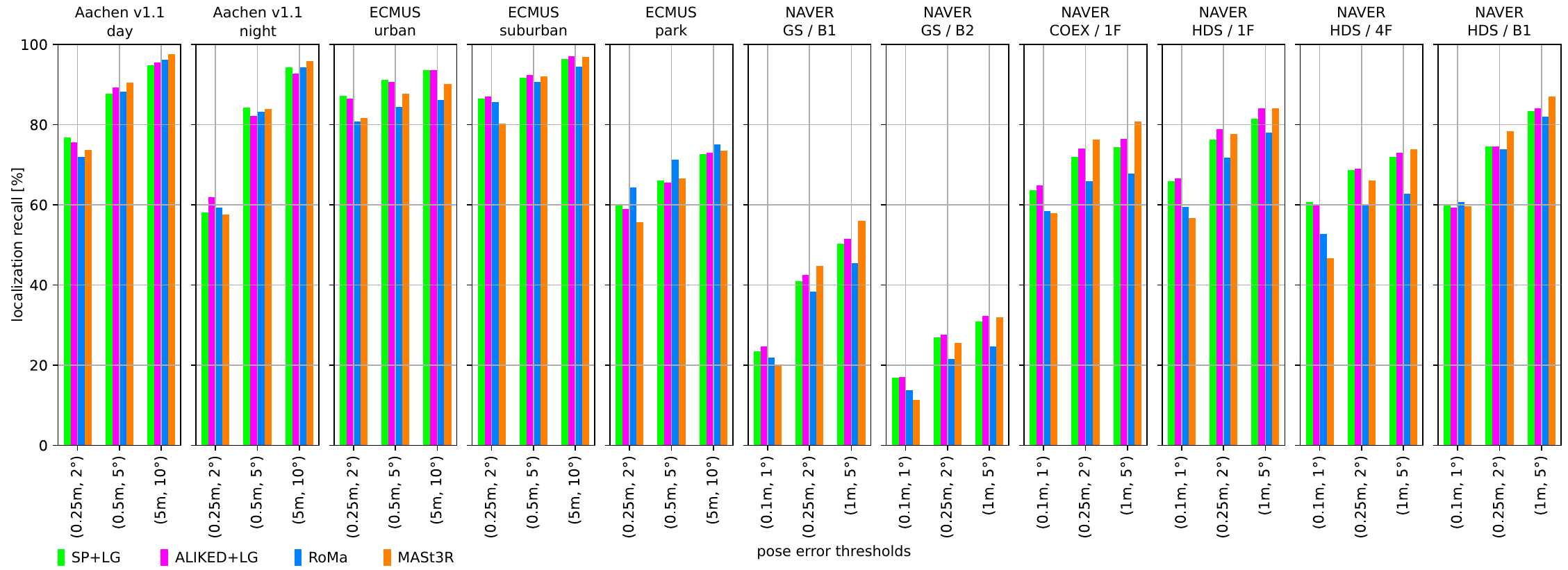}
    \vspace{-0.4cm}
    \caption{\emph{LazyLoc} localization results for different features. We report localization recalls (higher is better) on the Y-axis at multiple pose thresholds (X-axis). There is no type of feature that performs best in all scenes. However, the MASt3R matcher performs well in general.}
    \label{fig:recalls_lazyloc}
\end{figure*}

\begin{figure*}[!tb]
    \centering
    \includegraphics[width=1\linewidth, trim=0 0.1cm 0 0.1cm, clip]{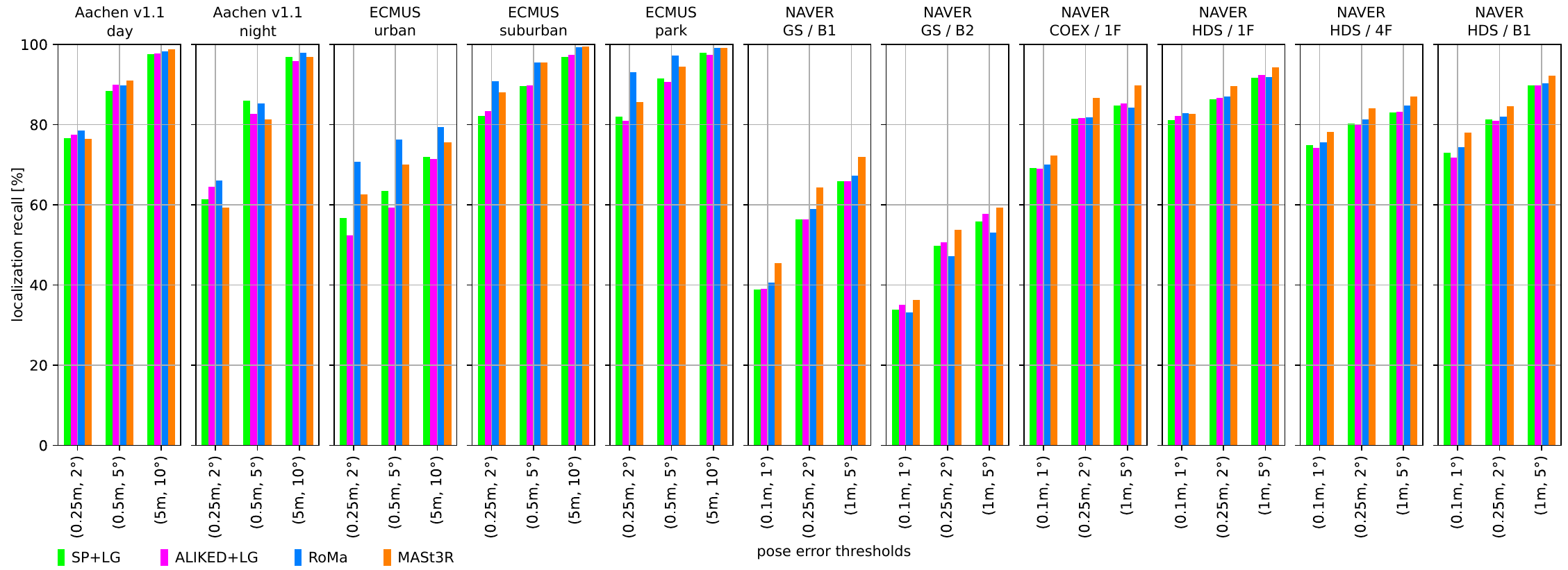}
    \vspace{-0.5cm}
    \caption{\emph{E5+1} localization results for different features. We report localization recalls (higher is better) on the Y-axis at multiple pose thresholds (X-axis). For the outdoor scenes, the best results are typically obtained with the RoMa matcher. For the indoor scenes, the MASt3R matcher performs best for the coarser thresholds.}
    \label{fig:recalls_e5+1}
\end{figure*}

\begin{figure*}[!tb]
    \centering
    \includegraphics[width=1\linewidth, trim=0 0.1cm 0 0.1cm, clip]{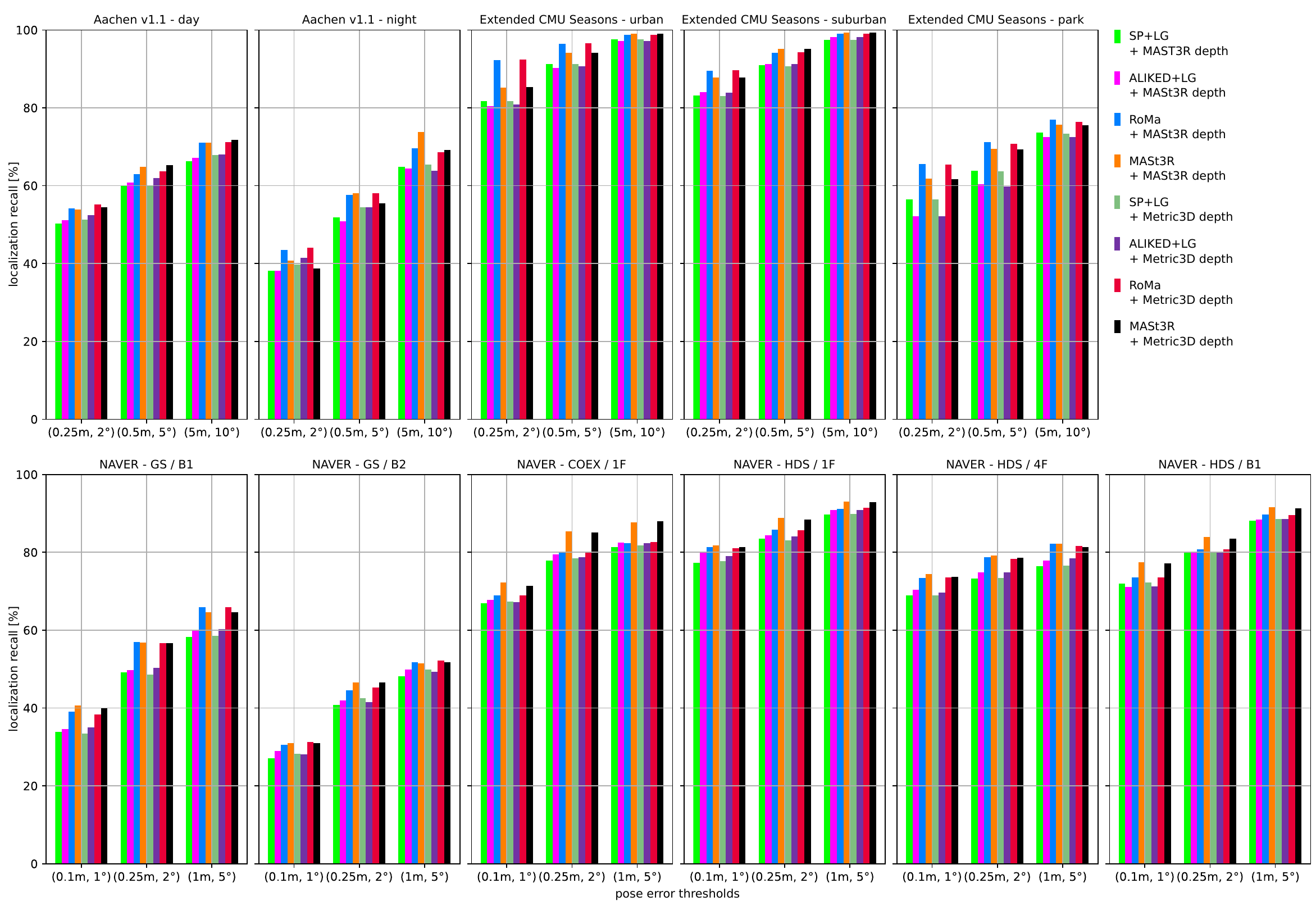}
    \vspace{-0.5cm}
    \caption{\emph{E3+1} localization results for different features and depth predictors. We report localization recalls (higher is better) on the Y-axis at multiple pose thresholds (X-axis). The choice of the depth predictor is not critical.}
    \label{fig:recalls_depth_genrelpose}
\end{figure*}

\begin{figure*}[!tb]
    \centering
    \includegraphics[width=1\linewidth, trim=0 0.1cm 0 0.1cm, clip]{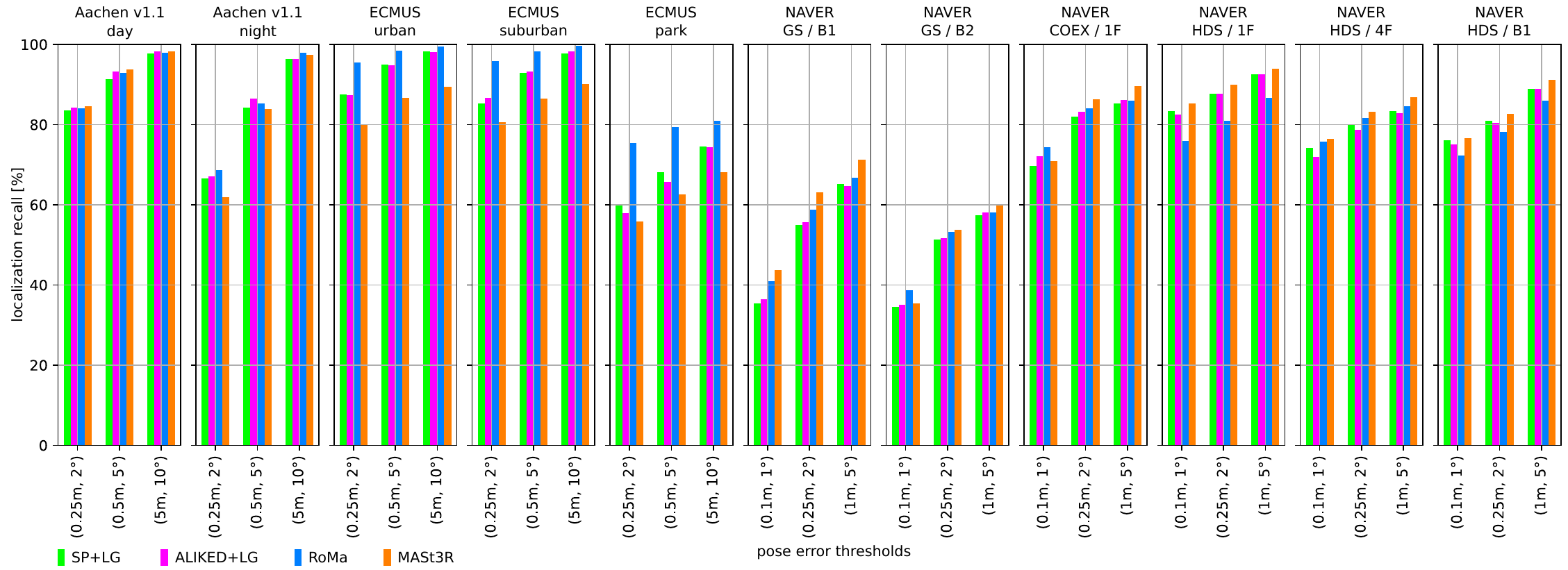}
    \vspace{-0.5cm}
    \caption{Localization results for the local 3D point triangulation from all retrieved images (\emph{Local triangulation - all}) for different features. We report localization recalls (higher is better) on the Y-axis at multiple pose thresholds (X-axis).}
    \label{fig:recalls_local_triang_all}
\end{figure*}

\begin{figure*}[!tb]
    \centering
    \includegraphics[width=1\linewidth, trim=0 0.1cm 0 0.1cm, clip]{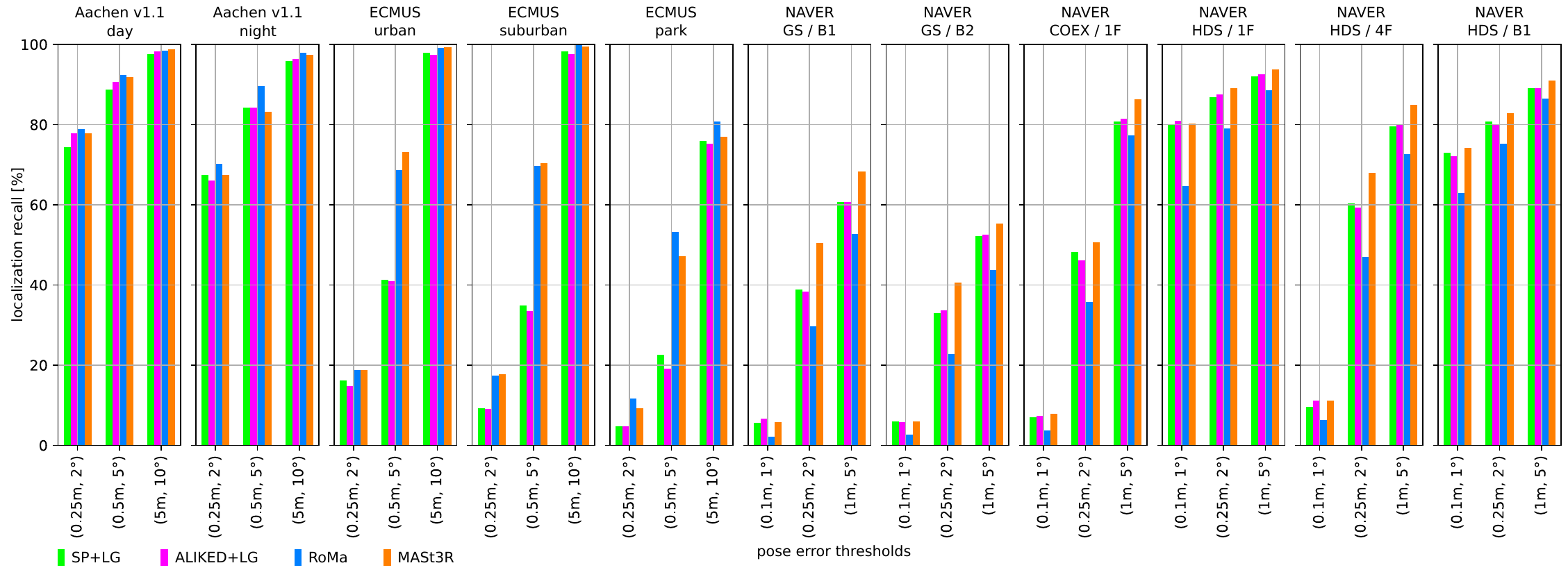}
    \vspace{-0.5cm}
    \caption{Localization results for the local 3D point triangulation from reference image pairs (\emph{Local triangulation - pairs}) for different features. We report localization recalls (higher is better) on the Y-axis at multiple pose thresholds (X-axis).}
    \label{fig:recalls_local_triang_pairs}
\end{figure*}

In a first set of experiments, we evaluate the impact of certain design choices (typically the type of features used and the type of depth maps used when applicable) on the performance of the various structureless localization methods. 
Sec.~\ref{sec:experiments:comparisons_structureless} then compares the best-performing versions of the different structureless methods with each other. 
Finally, Sec.~\ref{sec:experiments:comparisons_structurebased} compares structureless methods with structure-based methods.

\noindent \textbf{Pose triangulation.} 
We present the results for \emph{Ess. mat.} using the 5Pt solver in Fig.~\ref{fig:recalls_rel_pose_triang} and using the 3Pt+depth solver in Fig~\ref{fig:recalls_depth_relpose}. 
As can be seen from Fig~\ref{fig:recalls_depth_relpose}, the choice of the depth predictor does not seem to be critical, with both predictors performing similarly well.
Note that for the \emph{Ess. mat.} approach, the scale of the depth maps is not important as we use relative poses up to the scale of the translation.
As can be seen from both figures, the RoMa matcher performs best for the outdoor scenes, while the MASt3R matcher typically yields the best accuracy in indoor scenes. 

Fig.~\ref{fig:recalls_lazyloc} shows results for \emph{LazyLoc} when using different features. 
In contrast to the \emph{Ess. mat.} approaches, there is no feature type that performs best in all scenes (or indoor or outdoor scenes). 
However, the MASt3R matcher gives good results in general and seems to be the feature of choice.

\noindent \textbf{Semi-generalized relative pose estimation.} 
Figures~\ref{fig:recalls_e5+1} and~\ref{fig:recalls_depth_genrelpose} show results for the \emph{E5+1} respectively the \emph{E3+1} approach. 
For \emph{E5+1}, the dense RoMA matcher often performs best in outdoor scenes while the dense MASt3R matcher performs best indoors. 
For \emph{E3+1}, the RoMA matcher performs best on the Extended CMU Seasons dataset, while the MASt3R matcher works similarly well or better in all other scenes. 
As for \emph{Ess. mat. (3Pt + depth)}, the choice of the monocular depth predictor is not critical. 
Note that, as for \emph{Ess. mat. (3Pt + depth)}, the scale of the predicted depth maps is not important as the additional point correspondence with a second database image is used to recover the scale of the translation. 

\noindent \textbf{SfM on the fly.}
The results for the two local 3D point triangulation methods are shown in Fig.~\ref{fig:recalls_local_triang_all} respectively Fig~\ref{fig:recalls_local_triang_pairs}.
For the \emph{Local triangulation - all} approach (Fig.~\ref{fig:recalls_local_triang_all}), we observe that all features (except MASt3R) perform similarly well on Aachen Day-Night. 
For the Extended CMU Seasons dataset, RoMA clearly provides the best results, while MASt3R is typically the best for the indoor scenes. 
Interestingly, both SuperPoint and ALIKED outperform MASt3R on the Extended CMU dataset. 
In contrast, for \emph{Local triangulation - pairs} (Fig.~\ref{fig:recalls_local_triang_pairs}), both RoMa and MASt3R perform similarly to or better than SuperPoint and ALIKED. 
Interestingly, MASt3R performs best on Extended CMU when using pairs, while performing worse when using all images for triangulation. 
We speculate that some of the keypoint positions of MASt3R can be rather noisy. 
This negatively impacts the accuracy of the triangulated 3D points when using all images. 
When using pairs, there is a chance to obtain more accurate point positions by selecting two database images with consistent keypoints. 
In contrast, the other features benefit from using all images. 

\begin{table*}[!tb]
\centering
\small
\begin{tabular}{l l l l}
method &matching & day & night \\ \hline
MASt3R pose align~\cite{dust3r_cvpr24, MASt3R_eccv24} &MASt3R & 9.6 / 39.6 / 88.6 & 9.4 / \textbf{37.2} / \textbf{88.0}\\
MASt3R depth~\cite{dust3r_cvpr24,MASt3R_eccv24} + P3P~\cite{Persson2018ECCV} &MASt3R & 0.1 / 1.1 / 37.7 & 0.0 / 0.5 / 33.0 \\
Ess. mat. (MASt3R poses)~\cite{Zhou2020ICRA, dust3r_cvpr24,MASt3R_eccv24} & MASt3R & \textbf{23.2} / \textbf{49.0} / \textbf{91.0}& \textbf{12.0} / 28.3 / 82.7\\
Reloc3r~\cite{dong2024reloc3r} &- & 5.2 / 14.2 / 63.0 & 2.1 / 6.8 / 54.5 \\
\end{tabular}
\caption{Comparison of regression-based methods on Aachen Day-Night v1.1~\cite{Zhang2020ARXIV, Sattler2018CVPR, Sattler2012BMVC}. We use the top 10 images retrieved using the  EigenPlaces~\cite{Berton_2023_EigenPlaces} image-level descriptor. We report localization recalls (higher is better) at the pose thresholds of (0.25m, 2°) / (0.5m, 5°) / (5m, 10°).}
\label{tab:regres_methods}
\end{table*}

\noindent \textbf{Relative pose regression.} 
Tab.~\ref{tab:regres_methods} shows results for the four regressor-based methods on the Aachen Day-Night v1.1~\cite{Zhang2020ARXIV,Sattler2018CVPR, Sattler2012BMVC} dataset.
Compared to all previously evaluated approaches, all variants of pose regression-based methods perform significantly worse. 
The \emph{Ess. mat. (MASt3R poses)} performs the best among these methods, but is still significantly less accurate than \emph{Ess. mat. (5Pt)}. 
Even though the MASt3R depth regressor was trained to predict metric depth maps, our experiments show that the depths are far from metric. 
Thus, using them for lifting 2D-2D correspondence to 2D-3D matches, followed by P3P-RANSAC-based pose estimation, results in very inaccurate poses (see ~\emph{MASt3R depth + P3P} in Tab.~\ref{tab:regres_methods}). 
Our \emph{MASt3R pose align} pipeline achieves better results than \emph{Reloc3r}, but is still less accurate than \emph{Ess. mat. (MASt3R poses)}. 

Given the inaccurate pose estimates observed for the Aachen dataset, coupled with long run-times, we did not evaluate the pose regression-based approaches on other datasets. 

\begin{table*}[tb]
\centering
\small
\begin{tabular}{l r@{\hskip -2pt} c@{\hskip -2pt} l}
method & & time & \\ \hline
\textbf{Ess. mat. (5Pt)} & \textbf{1622.71 ms} & \textbf{/} & \textbf{query} \\
\quad ess. matrix estimation (5Pt) & \quad 30.37 ms & / & query-ref. pair \\
\quad pose estimation (cam. triangulation) & \quad 0.34 ms & / & ref. pair sample \\
\quad local optimization & \quad 32.05 ms & / & query \\ \hline

\textbf{Ess. mat. (3Pt+depth)} & \textbf{1552.83 ms} & \textbf{/} & \textbf{query} \\
\quad ess. matrix estimation (3Pt+depth) & \quad 10.86 ms & / & query-ref. pair \\
\quad pose estimation (cam. triangulation) & \quad 0.29 ms & / & ref. pair sample \\
\quad local optimization & \quad 30.67 ms & / & query \\ \hline

\textbf{LazyLoc} & \textbf{101.33 ms} & \textbf{/} & \textbf{query} \\
\quad graph preparation & \quad 19.45 ms & / & query \\
\quad pose averaging & \quad 4.31 ms & / & query \\
\quad bundle adjustment & \quad 34.71 ms & / & query \\ \hline

\textbf{E5+1} & \textbf{265.09 ms} & \textbf{/} & \textbf{query} \\
\quad pose estimation (E5+1 solver) & \quad 237.52 ms & / & query \\ \hline

\textbf{E3+1} & \textbf{477.42 ms} & \textbf{/} & \textbf{query} \\
\quad pose estimation (E3+1 solver) & \quad 351.55 ms & / & query \\ \hline

\textbf{Local triang. - all} & \textbf{2908.96 ms} & \textbf{/} & \textbf{query} \\
\quad point triangulation (with RANSAC) & \quad 3.49 ms & / & 3D point \\
\quad pose estimation (P3P solver) & \quad 14.01 ms & / & query \\ \hline

\textbf{Local triang. - pairs} & \textbf{1527.97 ms} & \textbf{/} & \textbf{query} \\
\quad point triangulation (from image pairs) & \quad 0.05 ms & / & 3D point \\
\quad pose estimation (P3P solver) & \quad 5.86 ms & / & ref. pair sample \\
\end{tabular}
\caption{Average runtimes of the evaluated methods, measured on Aachen Day-Night v1.1~\cite{Zhang2020ARXIV, Sattler2018CVPR, Sattler2012BMVC}. The experiments were performed on an Intel Core i7-9750H (2.60 GHz) CPU, using pre-computed SuperPoint~\cite{DeTone2017SuperPointSI} features matched by LightGlue~\cite{lindenberger2023lightglue}.}
\label{tab:times}
\end{table*}

\noindent \textbf{Discussion.} 
For most of the methods, RoMa~\cite{edstedt2024roma} (with its outdoor model) usually performs the best for outdoor data sets (especially the Extended CMU Seasons dataset), while MASt3R~\cite{MASt3R_eccv24} typically performs better for indoor data sets. 
For indoor datasets, we observed that using RoMa's outdoor model instead of its indoor model gives on average similar results.

Sparse features (SuperPoint and ALIKED) can achieve a similar or even better accuracy than dense matchers for some methods and scenes (\eg, \emph{LazyLoc} on most scenes and \emph{Local triangulation} methods for the Aachen dataset). 
For other methods and certain scenes, they perform significantly worse, especially for \emph{Local triangulation - pairs} on the Extended CMU Seasons dataset). 
Overall, dense matching methods often perform similar or better than sparse features. 
Still, they are not always the best choice.

The two depth estimators used for \emph{Ess. mat. (3Pt+depth)} and \emph{E3+1} lead to very similar pose accuracies, despite that we can see in Fig.~\ref{fig:depth_maps}. 
As observed above, both monocular depth estimators differ in the accuracy of their scale estimates. 
However, as also discussed above, the scale of the depth estimates is not used by either of the two methods. 
Still, Fig.~\ref{fig:depth_maps} shows clear ocal depth errors present in the Metric3D depth maps~\cite{yin2023metric, Hu2024Metric3DVA}. 
This is likely due to two reasons: 
By applying the 3Pt solver~\cite{Ding2025RePoseDER} inside RANSAC, we can ignore regions with inaccurate depth estimates. 
At the same time, the initial poses obtained via the 3Pt solver are later refined. 
The refinement does not use the depth maps, thus allowing us to better handle inaccuracies. 

\subsection{Comparing Structureless Localization \\Approaches}
\label{sec:experiments:comparisons_structureless}

\begin{table*}[!tb]
\centering
\footnotesize
\begin{tabular}{l l ll l}
method & matching  &depth& day & night \\ \hline
Ess. mat. (5Pt)~\cite{Zhou2020ICRA} & RoMa  &-&  59.1 / 80.0 / 97.9 &  52.4 / 78.0 / 97.4\\
Ess. mat. (3Pt+depth) &  RoMa  &Metric3D & 66.7 / 86.8 / \textbf{98.4}& 63.4 / 87.4 / \textbf{97.9}\\
LazyLoc~\cite{dong2023lazy}& MASt3R  &-& 73.7 / 90.5 / 97.5 & 57.6 / 83.8 / 95.8\\ \hline
E5+1 & RoMa  &-& 78.4 / 89.8 / 97.8 & 65.4 / 84.8 / \textbf{97.9}\\
E3+1 &  RoMa  &Metric3D &  54.2 / 63.0 / 70.1&  46.6 / 60.7 / 71.7\\ \hline
Local triang. - all & RoMa  &-&  \textbf{84.0} / \textbf{92.8} / 97.9&  68.6 / 85.3 / \textbf{97.9}\\
Local triang. - pairs & RoMa  &-&  78.8 / 92.4 / \textbf{98.4}&  \textbf{70.2} / \textbf{89.5} / \textbf{97.9}\\
\end{tabular}
\caption{The best performing setup (matching method and depth map source) for each method evaluated on Aachen Day-Night v1.1~\cite{Zhang2020ARXIV, Sattler2018CVPR, Sattler2012BMVC}. We use the top 10 images retrieved using the  EigenPlaces~\cite{Berton_2023_EigenPlaces} image-level descriptor. We report localization recalls (higher is better) at pose thresholds of (0.25m, 2°) / (0.5m, 5°) / (5m, 10°).}
\label{tab:benchmark_aachen}
\end{table*}

\begin{table*}[!tb]
\vspace{0.3cm}
\centering
\footnotesize
\begin{tabular}{l l ll l l}
method & matching  &depth& urban & suburban & park \\ \hline
Ess. mat. (5Pt)~\cite{Zhou2020ICRA} & RoMa  &-&  81.2 / 89.7 / 97.0&  84.9 / 91.7 / 97.9&  65.3 / 73.0 / 78.8\\
Ess. mat. (3Pt+depth) &  RoMa  &MASt3R &  85.3 / 92.4 / 97.8&  87.2 / 92.5 / 97.9& 67.7 / 74.2 / 79.0\\
LazyLoc~\cite{dong2023lazy}& SP+LG  &-& 87.1 / 91.1 / 93.6& 86.4 / 91.7 / 96.4& 60.1 / 66.0 / 72.7\\ \hline
E5+1 & RoMa  &-& 93.0 / 97.2 / 99.1 & 90.8 / 95.4 / 99.2 & 70.7 / 76.3 / 79.3 \\
E3+1 &  RoMa  &MASt3R &  92.3 / 96.5 / 98.8&  89.5 / 94.1 / 99.0& 65.5 / 71.2 / 77.0\\ \hline
Local triang. - all & RoMa  &-& \textbf{95.5} / \textbf{98.5} / \textbf{99.5}& \textbf{95.8} / \textbf{98.3} / 99.7& \textbf{75.4} / \textbf{79.3} / \textbf{81.0}\\
Local triang. - pairs & RoMa  &-&  18.8 / 68.7 / 99.1&  17.4 / 69.7 / \textbf{99.8}&  11.6 / 53.3 / 80.7\\
\end{tabular}
\caption{The best performing setup (matching method and depth map source) for each method. Evaluated on Extended CMU Seasons~\cite{Sattler2018CVPR, Badino2011}. We use the top 10 images retrieved using the  EigenPlaces~\cite{Berton_2023_EigenPlaces} image-level descriptor. We report localization recalls (higher is better) at the pose thresholds of (0.25m, 2°) / (0.5m, 5°) / (5m, 10°).}
\label{tab:benchmark_cmu}
\end{table*}

\begin{table*}[!tb]
\vspace{0.3cm}
\centering
\footnotesize
\begin{tabular}{l l ll l l}
method & matching  &depth& GS B1 & GS B2 & COEX 1F \\ \hline
Ess. mat. (5Pt)~\cite{Zhou2020ICRA} & MASt3R&-&  26.7 / 55.6 / 67.6&  21.7 / 43.1 / 52.0&  45.6 / 78.1 / 87.8\\
Ess. mat. (3Pt+depth) &  MASt3R&MASt3R&  27.2 / 55.1 / 68.1&  21.7 / 44.3 / 52.8& 47.4 / 79.7 / 87.8\\
LazyLoc~\cite{dong2023lazy}& ALIKED+LG&-& 24.6 / 42.5 / 51.5& 17.0 / 27.6 / 32.3& 64.9 / 74.0 / 76.4\\ \hline
E5+1 & MASt3R&-& \textbf{45.4} / \textbf{64.3} / \textbf{72.0}& \textbf{36.2} / \textbf{53.8} / 59.3& \textbf{72.3} / \textbf{86.7} / \textbf{89.7}\\
E3+1 &  MASt3R&MASt3R&  40.7 / 56.8 / 64.6&  31.0 / 46.5 / 51.5&  \textbf{72.3} / 85.4 / 87.7\\ \hline
Local triang. - all & MASt3R&-&  43.7 / 63.1 / 71.3&  35.3 / \textbf{53.8} / \textbf{59.8}&  70.9 / 86.3 / 89.6\\
Local triang. - pairs & MASt3R&-&  5.7 / 50.5 / 68.3&  6.0 / 40.5 / 55.3&  7.9 / 50.7 / 86.3\\
\end{tabular}
\caption{Benchmark on NAVER indoor localization datasets~\cite{Lee2021LargescaleLD} Gangnam Station (GS) and COEX scenes. We use the top 10 images retrieved using the  EigenPlaces~\cite{Berton_2023_EigenPlaces} image-level descriptor. We report localization recalls (higher is better) at the pose thresholds of (0.1m, 1°) / (0.25m, 2°) / (1m, 5°).}
\label{tab:benchmark_naver_gs}
\end{table*}

\begin{table*}[!tb]
\vspace{0.3cm}
\centering
\footnotesize
\begin{tabular}{l l l l l l}
method & matching  &depth& HDS 1F & HDS 4F & HDS B1 \\ \hline
Ess. mat. (5Pt)~\cite{Zhou2020ICRA} & MASt3R&-&  64.3 / 81.1 / 91.0&  53.4 / 74.6 / 82.3&  66.3 / 80.8 / 90.5\\
Ess. mat. (3Pt+depth) &  MASt3R&MASt3R&  66.6 / 83.1 / 91.0&  55.2 / 75.4 / 83.3& 67.0 / 80.4 / 91.2\\
LazyLoc~\cite{dong2023lazy}& ALIKED+LG&-& 66.6 / 78.9 / 84.1& 59.8 / 68.9 / 72.9& 59.2 / 74.6 / 84.0\\ \hline
E5+1 & MASt3R&-& 82.6 / \textbf{89.5} / \textbf{94.2}& \textbf{78.2} / \textbf{84.1} / \textbf{86.9}& \textbf{78.0} / \textbf{84.5} / \textbf{92.2}\\
E3+1 &  MASt3R&MASt3R&  81.7 / 88.8 / 93.0&  74.4 / 79.2 / 82.2&  77.4 / 84.0 / 91.6\\ \hline
Local triang. - all & MASt3R&-&  \textbf{85.3} / 89.9 / 93.9&  76.4 / 83.2 / 86.8&  76.6 / 82.6 / 91.1\\
Local triang. - pairs & MASt3R&-&  80.3 / 89.0 / 93.8&  11.2 / 68.0 / 84.9&  74.1 / 82.8 / 90.9\\
\end{tabular}
\caption{Benchmark on NAVER indoor localization datasets~\cite{Lee2021LargescaleLD} Hyundai Department Store (HDS) scenes. We use the top 10 images retrieved using the  EigenPlaces~\cite{Berton_2023_EigenPlaces} image-level descriptor. We report localization recalls (higher is better) at the pose thresholds of (0.1m, 1°) / (0.25m, 2°) / (1m, 5°).}
\label{tab:benchmark_naver_hds}
\end{table*}

In the next set of experiments, we compare the different structureless localization approaches with each other. 
As discussed, we exclude pose regression-based approaches. 
For each method, we selected the best-performing setup (type of features and depth estimator) per dataset. 

Tab.~\ref{tab:benchmark_aachen} shows results for the Aachen Day-Night v1.1~\cite{Zhang2020ARXIV, Sattler2018CVPR, Sattler2012BMVC} dataset, Tab.~\ref{tab:benchmark_cmu} shows results for the Extended CMU Seasons~\cite{Sattler2018CVPR, Badino2011} dataset, and Tab.~\ref{tab:benchmark_naver_gs} and Tab.~\ref{tab:benchmark_naver_hds} show results for the NAVER indoor localization~\cite{Lee2021LargescaleLD} datasets. 
In addition, 
Tab.~\ref{tab:times} reports the average run times of the localization methods on the Aachen Day-Night v1.1~\cite{Zhang2020ARXIV, Sattler2018CVPR, Sattler2012BMVC} dataset.

In terms of pose accuracy, methods that rely on more extensive geometric reasoning perform in general better: 
\emph{Ess. mat. (5Pt)}, which uses pairwise relative poses obtained from 2D-2D matches, performs consistently worse than \emph{Ess. mat. (3Pt+depth)}, which also uses depth maps for relative pose estimation. 
\emph{LazyLoc} uses motion averaging and refinement based on 2D-3D matches, which can further boost performance. 
Interestingly, \emph{LazyLoc} performs significantly worse than \emph{Ess. mat. (3Pt+depth)} in some scenes (Aachen Night, Extended CMU Seasons park, NAVER GS B1 \& B2, NAVER HDS 1F \& B1). 
In some of these scenes, even \emph{Ess. mat. (5Pt)} outperforms \emph{LazyLoc}, indicating that the additional refinement steps might not be effective under all conditions. 
The results of the pipeline using the 5Pt algorithm with rotation averaging and camera center triangulation (\emph{Ess. mat. - 5Pt}) are in line with the results reported in the literature on smaller datasets~\cite{dong2023lazy, arnold2022map}.

\emph{Ess. mat.} and \emph{LazyLoc} first compute pairwise relative pose estimates with the retrieved database images, which are then fused into the final query pose prediction. 
In contrast, \emph{E5+1} and \emph{E3+1} directly compute the query pose \wrt multiple database images, which further improves performance. 
Unlike for \emph{Ess. mat.}, using depth maps as part of the solver (\ie, using the E3+1 instead of the E5+1 solver) can lead to inferior results. 
However, this behavior seems scene-dependent: 
The accuracy drop is most notable for the Aachen dataset, where the scene structure is often tens of meters away from the query image. 
In such a scenario, inaccuracies in the depth maps propagate to large pose errors.
An interesting direction for future work is thus to automatically decide when to use monocular depth map predictions.

\begin{figure*}[tb]
    \centering
    \includegraphics[width=1\linewidth]{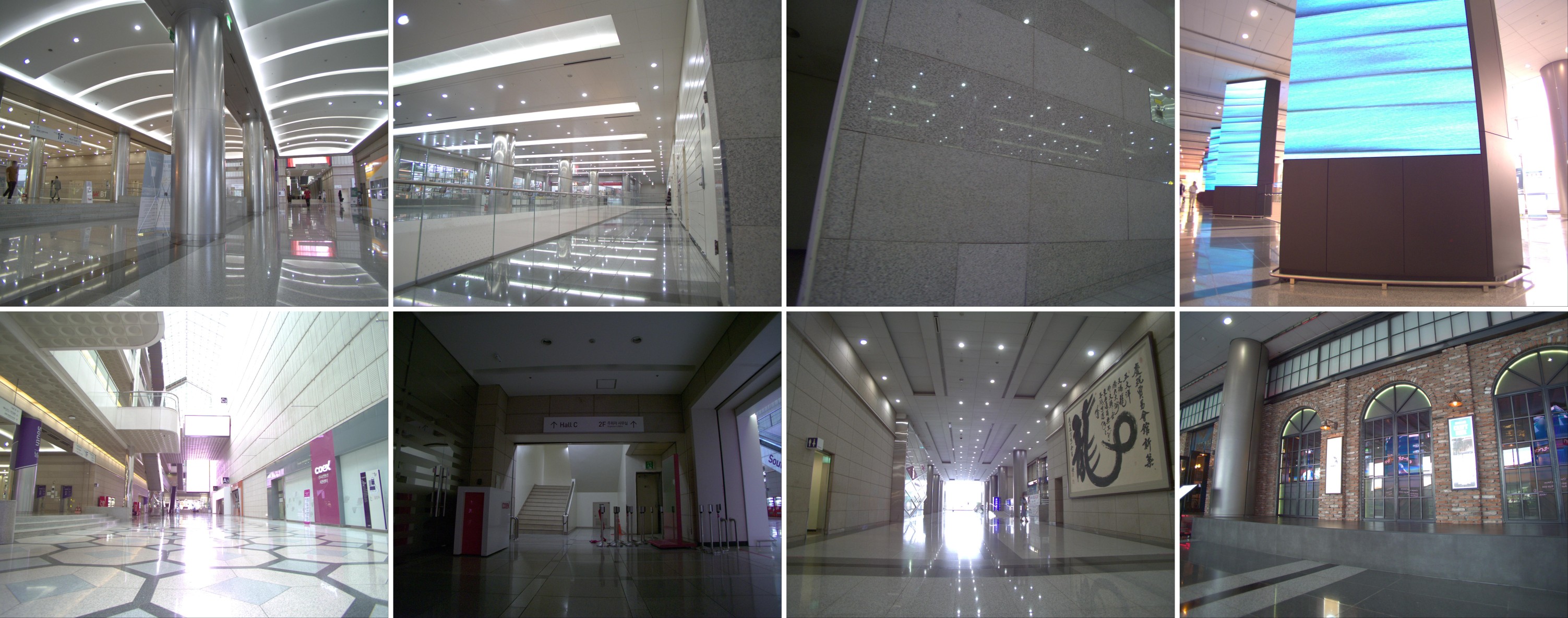}
    \vspace{-0.5cm}
    \caption{Comparison of the query images with high camera position error (top row) and the images with low camera position error (bottom row) in NAVER indoor localization dataset~\cite{Lee2021LargescaleLD} COEX 1F scene. The former often contain repetitive patterns or other structures that complicate 3D point triangulation.}
    \label{fig:naver_coex_examples}
\end{figure*}

Compared to \emph{E5+1} and \emph{E3+1}, \emph{Local triangulation} methods generate a local 3D model of the scene and use it to estimate the query pose. 
When using all pairs (\emph{Local triangulation - all}), this approach performs the best on the outdoor scenes, while being slightly less accurate than \emph{E5+1} on the indoor scenes. 
We observe a significant drop in pose accuracy when using only image pairs (\emph{Local triangulation - pairs}) on the Extended CMU dataset and in some indoor scenes (Gangnam Station B1 and B2, COEX 1F, and Hyundai Department Store 4F), which might be caused by narrow corridors, view occlusion by pedestrians, repetitive geometric patterns and textures and reflective surfaces (see Fig.~\ref{fig:naver_coex_examples}), making the triangulation from the image pairs more challenging compared to using all retrieved images.

In terms of pose accuracy, the \emph{Local triangulation - all} and \emph{E5+1} approaches are clearly the best choices. 
Taking run-times into account (\cf Tab.~\ref{tab:times}), the \emph{E5+1} offers a better trade-off between pose accuracy and run-time. 
When run-times are a major concern, \emph{LazyLoc} is a good alternative to these other two methods as it provides the fastest run-times while achieving reasonable pose accuracy on the outdoor datasets. 

\subsection{Comparison with Structure-based \\Methods}
\label{sec:experiments:comparisons_structurebased}

\begin{table*}[tb]
\centering
\small
\begin{tabular}{l ll l l}
method &   matching & top-k & day & night \\ \hline
\multirow{2}{*}{Hloc~\cite{sarlin2019coarse, sarlin2020superglue}} &  \multirow{2}{*}{SP+LG}&20 & \textbf{88.1} / \textbf{95.4} / 99.0& 71.7 / 89.0 / 97.9 \\
 &  &10& 87.0 / 94.8 / 98.5&70.2 / 87.4 / 97.4\\ \hline
\multirow{2}{*}{MeshLoc~\cite{Panek2022ECCV}} &   \multirow{2}{*}{SP+LG}&20& 85.9 / 93.6 / 98.8 & 70.2 / 87.4 / 97.9 \\
 &  &10& 84.2 / 92.5 / 98.5&70.2 / 85.9 / 96.9\\  \hline
MASt3R vis. loc.~\cite{MASt3R_eccv24} & R2D2 / MASt3R&20& 83.4/95.3/\textbf{99.4}& \textbf{76.4}/\textbf{91.6}/\textbf{100}\\ \hline \hline
\multirow{2}{*}{LazyLoc~\cite{dong2023lazy}} &   \multirow{2}{*}{SP+LG}&20& 77.7 / 88.8 / 94.3 & 58.1 / 84.3 / 94.2 \\
 &  &10& 76.8 / 87.7 / 94.7&58.1 / 84.3 / 94.2\\ \hline
\multirow{2}{*}{E5+1} &   \multirow{2}{*}{SP+LG}&20& 77.8 / 89.1 / 98.1 & 65.4 / 85.9 / 96.3 \\
 &  &10& 76.6 / 88.3 / 97.5&61.3 / 85.9 / 96.9\\ \hline
\multirow{2}{*}{Local triang. - all} &   \multirow{2}{*}{SP+LG}&20& 86.7 / 93.8 / 98.3 & 67.5 / 85.3 / 97.4 \\
 &  &10& 83.5 / 91.4 / 97.8&66.5 / 84.3 / 96.3\\
\end{tabular}
\caption{Comparison of the best structureless methods with state-of-the-art structure-based methods on Aachen Day-Night v1.1~\cite{Zhang2020ARXIV, Sattler2018CVPR, Sattler2012BMVC}. We use the top-10 and top-20 images retrieved using the  EigenPlaces~\cite{Berton_2023_EigenPlaces} image-level descriptor for all methods except \emph{MASt3R vis. loc.}. We report localization recalls (higher is better) at the pose thresholds of (0.25m, 2°) / (0.5m, 5°) / (5m, 10°). The first three method are structure-based approaches.}
\label{tab:sota_comparison}
\end{table*}

Our final experiment compares the best-performing structureless methods with  state-of-the-art structure-based methods:  
\emph{Hloc}~\cite{sarlin2019coarse, sarlin2020superglue} is a hierarchical structure-based pipeline using a SfM point cloud as a scene representation.
We evaluated Hloc with an SfM point cloud triangulated using the ground-truth reference camera poses. 
Each query image is matched against the top-10 respectively top-20 reference images retrieved using EigenPlaces~\cite{Berton_2023_EigenPlaces} image-level descriptors. 
Feature extraction and matching were performed using SuperPoint~\cite{DeTone2017SuperPointSI} and LightGlue~\cite{lindenberger2023lightglue}.
\emph{MeshLoc}~\cite{Panek2022ECCV} is a hierarchical pipeline that uses reference depth maps rendered from a triangular mesh to lift the 2D-2D query-reference correspondences to 2D-3D matches.
Our evaluation uses the AC-14 mesh, which in the original paper offered the best results.
\emph{MASt3R vis. loc.} is a hierarchical structure-based localization pipeline presented in~\cite{MASt3R_eccv24}.
It uses a precomputed 3D point cloud, triangulated using R2D2 features matches~\cite{revaud2019r2d2}, which is projected into the reference images in order to lift the 2D-2D correspondences established with the MASt3R matcher between the query and the retrieved reference images to 2D-3D correspondences.
The pose is then estimated using the P3P pose solver inside RANSAC.

Tab.~\ref{tab:sota_comparison} compares the structure-based methods with the best-performing structureless approaches from our previous experiments. 
As can be seen, Hloc holds the first position, with MeshLoc a little behind.
The best of the structureless methods is \emph{Local triang. - all}, being on par with MeshLoc on the day split of the dataset and a few percent behind on the night split.
The other two structureless methods \emph{LazyLoc} and \emph{E5+1} fall behind Hloc by approximately 10 percentage points on the smallest error thresholds, while being approximately 5 percentage points behind for the two larger error thresholds. 
Overall, we can see that structureless approaches can be competitive compared to structure-based methods. 
Coupled with the flexibility of their scene representation, structureless approaches thus represent an interesting alternative to the currently predominant structure-based methods.

\section{Conclusion}\label{sec:conclusion}
In this paper, we have provided a comprehensive overview and a detailed comparison of structureless visual localization approaches.
Through extensive experiments, we have compared different families of structureless approaches. 
Our results show that more extensive geometric reasoning typically leads to better performance, with the best results being obtained by building local SfM models on the fly. 
However, the best accuracy-run-time tradeoff is provided by methods based on semi-generalized relative pose estimation. 
Our experiments did not reveal a single best localization or matching method, but they can serve as a reference for the reader to select the method based on the use case. 
We also looked into regression-based methods, which, from our evaluation, still need to mature to achieve the accuracy of more classical approaches.
Compared to structure-based approaches, structureless methods are in general less accurate, but can perform comparably. 

\clearpage % Prevents large space between the "Conclusion" title and the rest of the text

\section*{Acknowledgments}
This project was supported by the Czech Science Foundation (GAČR) JUNIOR STAR Grant No. 22-23183M and EXPRO Grant No. 23-07973X, by the ERC Starting Grant DynAI (ERC-101043189) and by the Grant Agency of the Czech Technical University in Prague, grant No. SGS23/121/OHK3/2T/13.

\FloatBarrier  % Push all figures and tables above this line

\small{

}

\end{document}